# Bridging Data Barriers among Participants: Assessing the Potential of Geoenergy through Federated Learning

Weike Peng[a,b], Jiaxin Gao[a,c], Yuntian Chen[a*], Shengwei Wang[b,d*]

a. Ningbo Institute of Digital Twin, Eastern Institute of Technology, Ningbo, P. R. China

b. Department of Building Environment and Energy Engineering, The Hong Kong Polytechnic University, Kowloon, Hong Kong

c. School of Electronic Information and Electrical Engineering, Shanghai Jiao Tong University, Shanghai, P. R. China

d. Research Institute for Smart Energy, The Hong Kong Polytechnic University, Kowloon, Hong Kong

*Corresponding authors. E-mail: ychen@eitech.edu.cn (Dr. Yuntian Chen)

*Corresponding authors. E-mail: shengwei.wang@polyu.edu.hk (Prof. Shengwei Wang)

**Abstract**:

Machine learning algorithms emerge as a promising approach in energy fields, but its practical is hindered by data barriers, stemming from high collection costs and privacy concerns. This study introduces a novel federated learning (FL) framework based on XGBoost models, enabling safe collaborative modeling with accessible yet concealed data from multiple parties. Hyperparameter tuning of the models is achieved through Bayesian Optimization. To ascertain the merits of the proposed FL-XGBoost method, a comparative analysis is conducted between separate and centralized models to address a classical binary classification problem in geoenergy sector. The results reveal that the proposed FL framework strikes an optimal balance between privacy and accuracy. FL models demonstrate superior accuracy and generalization capabilities compared to separate models, particularly for participants with limited data or low correlation features and offers significant privacy benefits compared to centralized model. The aggregated optimization approach within the FL agreement proves effective in tuning hyperparameters. This study opens new avenues for assessing unconventional reservoirs through collaborative and privacy-preserving FL techniques.

**Keywords**: Data privacy and security; Federated learning; Geoenergy; Bayesian Optimization; Production estimation.

# 1 Introduction

Geoenergy, a term encompassing the diverse array of energy resources derived from the Earth, plays a pivotal role in powering our modern society. This includes both renewable and non-renewable resources, ranging from fossil fuels, such as coal, oil, and natural gas, to sustainable energy, such as geothermal energy and hydrogen energy [1, 2]. Faced with the dual pressures of increasing energy demand and climate change, the focus of scientists has been gradually shifting from traditional geoenergy to renewable geoenergy in recent decades. But a rapid and radical withdrawal from fossil fuels appears infeasible if aims to sustain economic growth and meet growing energy demand [3, 4]. Moreover, by-products and derivatives of fossil fuels also have been not replaced in a short period. [5, 6]. Undoubtedly, the demand for oil and gas remains substantial, as predicted by British Petroleum [7].

As one of the important geoenergy resources, unconventional reservoir resources gradually garnered much attention [8]. It is defined as accumulations where gas phases are tightly bound to the rock fabric by strong capillary forces [9, 10]. Prediction and evaluation of estimated potential productivity (EPP) for gas wells is a fundamental task, offering valuable insights for decision-makers in the early stages of unconventional resource exploration [11-13]. Due to varied geological characteristics, construction conditions, and operation techniques, the EPP of the wells located in similar or even the same area differs greatly from each other [14]. Traditionally, these studies are generally carried out through complex mathematical calculations and physical modeling. As the starting point for productivity prediction of horizontal wells, analytical formulas based on Darcy's percolation law were initially proposed by Merkulov and Borisov [15]. Subsequent studies developed more accurate formulations for productivity estimation, with the main improvements lying in the utilization of advanced mathematical-physical methods and numerous experiments for investigating the impacts of geological features, and the physical characteristics of gas [16-20]. Based on the above-mentioned methods, the exploration potential of the gas wells in unconventional resources can be estimated accurately. However, these traditional models also face various challenges, such as high calculation costs, historical fitting difficulties, poor generative ability, low prediction efficiency, and high result uncertainty [21-24].

Machine learning (ML) is an application of artificial intelligence (AI) that provides systems the ability to automatically learn and improve from experience without being explicitly programmed [25, 26]. In recent years, the advent of AI has propelled the widespread adoption of this method in geoenergy fields [27]. It was demonstrated in the current literature that ML models can almost perfectly reflect the complex mechanisms and correlations between variables and predicted goals [28]. Additionally, it exhibits stronger predictive capacity, higher calculation efficiency, and better generalization ability, compared with conventional EPP methods [29]. Deep Neural Networks (DNN) [14, 30, 31], Random Forests (RF) [32, 33], eXtreme Gradient Boosting (XGBoost) [21, 34, 35], Support Vector Machines (SVM) [34, 36-38], and other classical ML models are frequently employed to predict the productivity of oil-gas wells with excellent accuracy [39-43]. Unfortunately, the practical implementation of ML algorithms is impeded by two prominent challenges: privacy protection and data collection. Firstly, the privacy of data has become an unavoidable concern for data administrators, as they are no longer permitted to access data without the user's permission after the General Data Protection Regulations published by the European Union in 2018 [44]. Under this agreement, users possess absolute ownership of their data [45, 46]. Secondly, collecting real-world data is laborious and expensive [49], particularly in the energy fields, where the average cost of a gas well or wind turbine can reach hundreds of thousands or even billions of dollars [48, 49]. Market participants desire an effective holistic model based on shared information across departments, but privacy concerns make them hesitant to share their data with business competitors, falling into a dilemma.

Federated learning (FL) is a multiparty setup where clients from different agencies collaboratively build a shared model across various datasets through exchanging intermediate computational results such as gradients or parameters, instead of raw data. In this configuration, a central server or active party securely aggregates calculation results from selected nodes to train or update a global model, addressing critical issues such as data privacy, data security, and data access rights [50]. As public concern about privacy continues to grow, FL is flourishing across various industries including grids, IoT, defense, and finance. In literature [51], multi-horizon FL was utilized for probabilistic forecasting of nodal voltages in local energy communities. Wang et al. [52] proposed a privacy-preserving clustering method to analyze clients' behavior in smart grids. FL models have also been employed to monitor the components in nuclear power plants, significantly improving prediction performance and reducing overfitting issues. Article [53] introduced a FL-based model for detecting wind turbine blade icing, providing real-time anomaly detection. In the field of environment protection, Hu [54] designed a novel environmental monitoring framework based on FL, effectively solving the challenges of inconvenient interchangeable monitor data from different regions. Saputra et al. [55] exploited a FL framework to predict energy demand for electric vehicle networks with satisfactory results.

To sum up, FL provides a fresh impetus to the growth of ML when data is not directly available due to some constraints, and the infinite potential of dispersed datasets could be fully exploited through it. However, to the best of our knowledge, there is no related research investigating the application of FL in the petroleum industry. With the advent of the Internet of Things (IoT) era, the demand for a safe, effective, and accurate data-sharing approach in the petroleum field is increasing dramatically. Therefore, this paper aims to introduce the FL technique into this area to overcome data barriers between agencies and companies, facilitating the application of ML methods to solve practical engineering problems. The benefits of FL agreements will be evaluated and analyzed on an actual dataset in different scenarios.

The contributions of this study are three-fold:

(1) Two novel federated learning frameworks are proposed to tackle the challenges arising from data reluctance and the inability to share among contributors when predicting the potential productivity of gas wells.

(2) By achieving data availability while preserving data invisibility, this research effectively resolves the issues of data scarcity and privacy. The validation of these advancements is supported by real-world data obtained from two unconventional gas reservoirs.

(3) Taking practicality into consideration, homomorphic encryption and Bayesian Optimization are employed to enhance the performance of the FL frameworks.

## 2 Methodology

Data scarcity and privacy concerns create a series of barriers to the application of data-driven methods in the petroleum industry. To eliminate these barriers, this section outlines the step-by-step workflow of the XGBoost model under the proposed FL agreement for estimating the production potential of gas wells in unconventional reservoirs. Firstly, a robust classifier generated by the XGBoost model is introduced to determine whether a gas well is worth developing. Next, a flexible and effective FL framework combined with the XGBoost classifiers (FL-XGBoost) is established to homogenize and safely consolidate data from disparate sources. Thirdly, partially homomorphic encryption is used to perform mathematical operations on encrypted data in various scenarios, leveraging its high efficiency and reliable security proof. Besides, the overall performance of the proposed strategy is compared and discussed based on the selected evaluation indexes. Finally, the Bayesian Optimization algorithm is applied to enhance the overall performance of the models through hyperparameter tuning. A detailed illustration of each of these steps (shown in Figure. 1) is explained as follows:

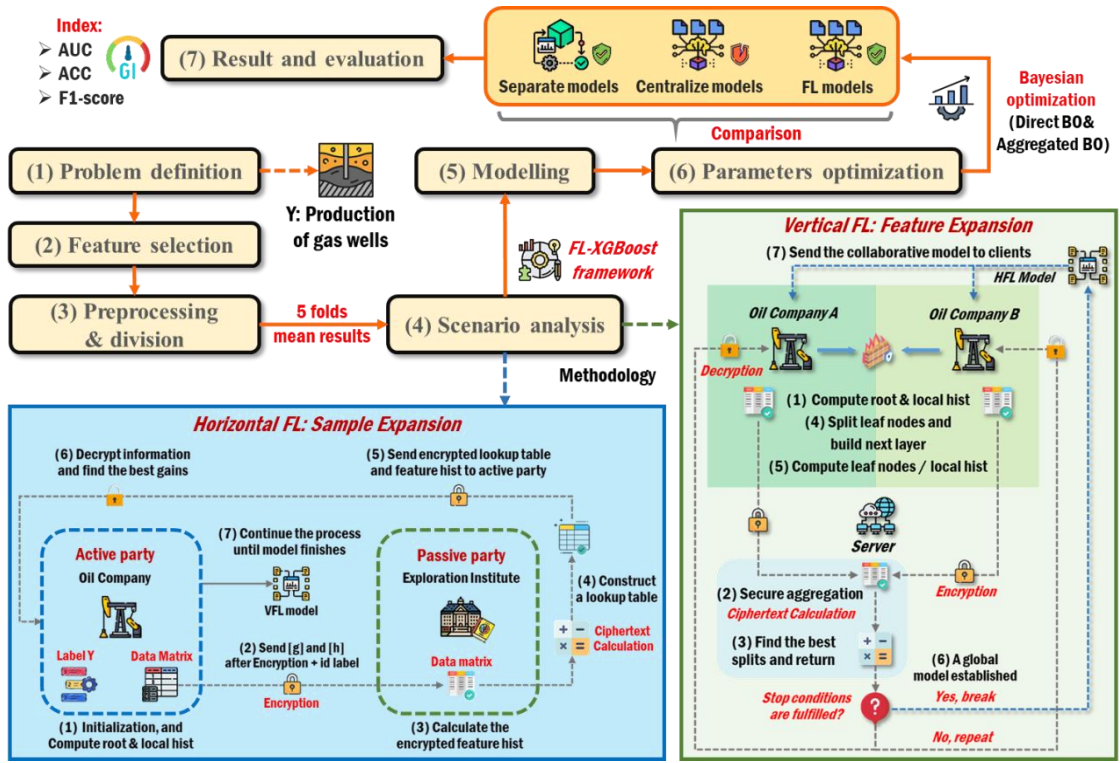

Figure. 1. Step-by-step workflow of the proposed federated learning approach.

## 2.1 Extreme Gradient Boosting

Extreme Gradient Boosting (XGBoost) offers a powerful gradient-boosting

framework to solve problems efficiently. It generates a robust classifier by iteratively updating the parameters of the former classifier to minimize the gradient of the loss function [56]. In this study, XGBoost algorithm serves as a valuable tool to assist administrators in making decisions regarding the development of gas wells.

The XGBoost object function $J$, consists of training loss function $L$ and regularization term $\Omega$, as shown in Eq. 1:

$$J(\theta) = L(\theta) + \Omega(\theta) \tag{1}$$

where $\theta$ indicates the parameters trained from a given set of data; $L$ is aimed to estimate the model's predictive ability, generally using square loss or logistic loss; $\Omega$ contains $L_1$ norm and $L_2$ norm, which measures the complexity of the model.

For a dataset $X \in R^{M \times N}$ with M sample and N features, the output of a XGBoost model $\hat{y}_i$, is averaged or voted by a collection $F$ of $k$ trees as follows [57]:

$$\hat{y}_i = \sum_{k=1}^{K} f_k(x_i), f_k \in F \tag{2}$$

where $f_k$ is a scoring function, corresponding to an independent tree structure and leaf weight in the Classification and Regression Tree (CART) algorithm, and $F$ is the space of CART.

In each boosting step, the output is updated by adding the score of one new tree at a time. The predication of XGBoost model at the $t$ time iteration is defined in Eq.3 [54]:

$$\hat{y}_i^{(t)} = \sum_{k=1}^{t} f_k(x_i) = \hat{y}_i^{(t-1)} + f_k(x_i) \tag{3}$$

Based on Eq.4 and Eq.5, the objective function at $t$ moment can be given as follows:

$$\begin{cases} J^{(t)} = \sum_{i=1}^{n} L(y_i, \hat{y}_i) + \sum_{k=1}^{t} \Omega(f_k) \\ J^{(t)} = \sum_{i=1}^{n} L(y_i, \hat{y}_i^{(t-1)} + f_k(x_i)) + \Omega(f_k) + constant \end{cases} \tag{4}$$

$$\Omega(f_k) = \gamma T + \frac{1}{2}\lambda \sum_{j=1}^{T} w_j^2 \tag{5}$$

where $n$ is the number of predictions of XGBoost model; $\gamma$ is the complexity parameter of each leaf; $T$ is the number of leaves; $\lambda$ is a parameter to scale the penalty; and $w$ is the vector of scores on leaves.

To quickly optimize the objective in XGBoost model, a second-order Taylor approximation can be used. When using mean square error as the loss function, the simplified objective function at step $t$ can be obtained by removing the constant terms.

$$\sum_{j=1}^{T} \left[ G_j w_j + \frac{1}{2} \left( H_j + \lambda \right) w_j^2 \right] + \gamma T \qquad (6)$$

where $g_i = \partial_{\hat{y}(t-1)} l\left(y_i, \hat{y}^{(t-1)}\right)$ and $h_i = \partial_{\hat{y}(t-1)}^2 l\left(y_i, \hat{y}^{(t-1)}\right)$ are first and second order gradient statistics on the loss function respectively [55].

The tree model starts with single leaf nodes which includes all samples. Then the node recursively splits the current samples into left and right subsets denoted by $I_L$ and $I_R$. The loss function after the split is:

$$\mathcal{L}_{split} = \frac{1}{2} \left[ \frac{(\Sigma_{i \in I_L} g_i)^2}{\Sigma_{i \in I_L} h_i + \lambda} + \frac{(\Sigma_{i \in I_R} g_i)^2}{\Sigma_{i \in I_R} h_i + \lambda} - \frac{(\Sigma_{i \in I_I} g_i)^2}{\Sigma_{i \in I_I} h_i + \lambda} \right] - \gamma \qquad (7)$$

where the best split is the one with the highest $\mathcal{L}_{split}$. The weight w of each leaf can be calculated in Eq.8.

$$\omega = -\frac{\Sigma_{i \in I_u} g_i}{\Sigma_{i \in I_u} h_i + \lambda} \qquad (8)$$

## 2.2 Federated learning

A vast amount of data from various sources supports well-informed decision-making and a more holistic view of hidden opportunities for each operator. However, the data availability of any party is limited. Collaboration among multiple parties is embraced as a powerful tool for data-driven methods to uncover and leverage new insights. Unfortunately, the participants frequently encounter challenges in data sharing, whether due to confidentiality or business reasons. When data is difficult or impossible to share, the ability to collaborate is hindered. Therefore, the adoption of FL is crucial to enable modeling with accessible yet concealed data from multiple parties.

According to the data distribution characteristics among participants, federated learning can be classified into three types: Horizontal Federated Learning (HFL), Vertical Federated Learning (VFL), and Federated Transfer Learning (FTL) [58]. Figure 2 illustrates the HFL and VFL frameworks in a two-party scenario.

Generally, HFL is commonly used when datasets share the same feature space across all devices, such as gas wells in two fields or customers from several petroleum companies. The HFL framework allows training datasets to be expanded in terms of sample numbers, potentially enhancing the generalization ability of ML models.

In contrast, VFL is feature-based federated learning that utilizes the distinct feature spaces from different datasets with the same ID of gas well to jointly train a global model. Participants in VFL agreements may have diverse professional backgrounds or play dissimilar roles in the industrial chain. For example, oil enterprises could collaborate with banks using VFL to build financial risk or credit assessment models for their customers. A broader range of data sources can be tremendously helpful for ML algorithms in providing more accurate results, particularly in cases where various factors may not have been adequately considered [59].

For $N$ distributed datasets $\{D_1, D_2, ..., D_N, \}$, each data owner locally develops $N$ prediction models $\{M_{local}^1, M_{local}^2, ..., M_{local}^N\}$ using their own data sources. Clearly, the generalizability of these separate models is probably poor. It is a wise choice for data-driven organizations that train a holistic model by consolidating their respective data. The collected datasets from all contributors $D_{SUM} = D_1 \cup D_2 \cup ... \cup D_N$ would be formed commonly through openly data sharing, and subsequently, a centralized model can be obtained. However, partial or full $D_{SUM}$ is inevitably exposed to some participants or a third party, under the huge risk and uncertainty of privacy leakage. FL frameworks enable safe collaborative modeling, and a secure global model would be obtained, in which process that data is accessible but invisible.

The cost of privacy protection, denoted as $C_{privacy}$, is defined as the expense incurred in safeguarding the privacy of participants in collaborative modeling. Clearly, utilizing the local data for separate modeling is secure, with its inherent privacy protection cost being zero. For FL models, the cost of privacy protection can be understood as the loss of performance incurred by introducing the FL framework, given the same available data sources. It can be described in Eq.9

$$C_{privacy}^{FL} = A_{open\,share} - A_{federated} \qquad (9)$$

where $A_{open\,share}$ is a performance score from the open share model and $A_{federated}$ is that from the FL model.

For centralized modeling with open data sharing, the cost of privacy protection can be defined as the difference in performance compared to separate local modeling under the same parameter settings.

$$C_{privacy}^{open\,share} = A_{open\,share} - A_{separate} \qquad (10)$$

where $A_{open\,share}$ is a performance score from the open share model, and $A_{separate}$ is that from the separate model.

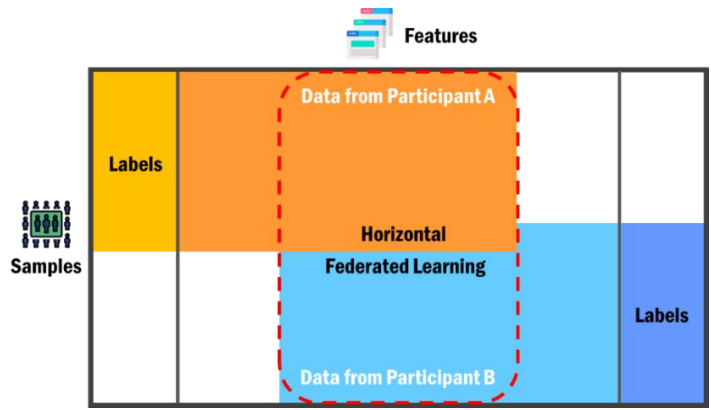

(a) Horizontal Federated Learning (Safe expansion in data samples)

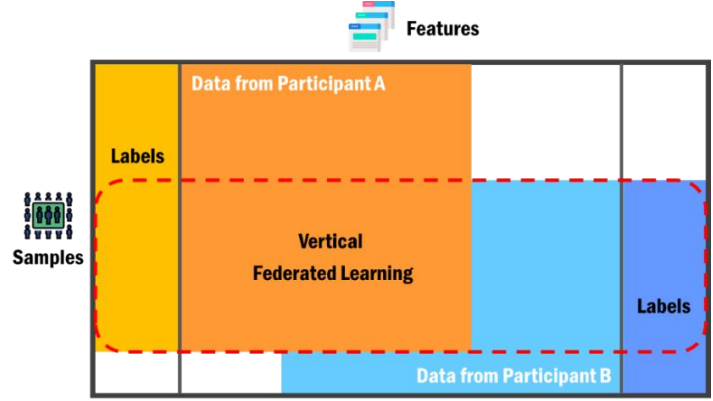

(b) Vertical Federated Learning (Safe expansion in data features)

Figure. 2. Categorization of federated learning.

## 2.3 Homomorphic Encryption

Throughout the stage of data aggregation, calculation, transmission, and storage, there is a possibility of inference attacks, resulting in the leakage of sensitive information. To address this issue, homomorphic encryption (HE) has gained recognition and extensive usage for secure aggregation and integrity verification in various application scenarios [60, 61]. According to the supported calculation types and support levels, HE can be classified into the following three types: Partially homomorphic encryption (PHE), Somewhat homomorphic encryption (SWHE), and Fully homomorphic encryption (FHE).

Among numerous homomorphic encryption schemes, PHE is the most commonly employed scheme in practical privacy computing due to its efficiency and complete security proof. Its prevalence can be attributed to the fact that, in specific scenarios, only one HE operation, such as addition, is necessary to fulfill certain functions. Hence, PHE is utilized in this article for conducting mathematical operations on encrypted data. The four key steps for PHE are as follows.

***Key generation***: Two large prime numbers $p$ and $q$ are randomly selected to calculate the Carmichael Function $\lambda(n)$ and Euler's Totient Modulus $\phi(n)$, represented as follow in Eq.11 and Eq.12. Importantly, $p$ and $q$ need to meet certain conditions, as shown in Eq. 12. Then, choose $g$ as a random integer where $g \in Z_{n^2}^*$, and the private element $\mu$ is obtained using Eq. 13.

$$\lambda(n) = lcm(p-1, q-1) \tag{11}$$

$$\phi(n) = (p-1) \times (q-1) \tag{12}$$

$$gcd\left(\phi(n), \lambda(n)\right) = 1 \tag{13}$$

where function *lcm* means the least common multiple of the input variable; $n$ is the product of the proposed prime numbers $p$ and $q$; function *gcd* means the greatest common divisor of the input variable [59].

$$\mu = (L(g^\lambda \bmod n^2))^{-1} \bmod n, \text{ where } L(u) = \frac{u-1}{n} \tag{14}$$

***Encryption***: For a given plaintext $m_i \in Z_n$, the encryption algorithm converts $m_i$ to the ciphertext as $C_i$. $r$ represents a given random number, where $0 \le r < n$ and $r \in Z_{n^2}^*$.

$$C_i = g^{m_i} \cdot r^n \bmod n^2 \tag{15}$$

***Secure aggregation***: when receiving ciphertexts $C_i$ from multiple parties, the aggregated ciphertext $C_{Ag}$ is calculated by Eq.15.

$$C_{Ag} = \prod_{i=1}^n C_i \bmod n^2 \tag{16}$$

***Decryption***: Upon receiving ciphertext $C_{Ag}$, the aggregated plaintext as $m$ can be computed by decryption algorithm as follows:

$$m = \sum_{i+1}^n m_i = R\left(C_i^\lambda \bmod n^2\right) \mu \bmod n \tag{17}$$

Noting:

(1) The public encryption key is $(n, g)$.

(2) The private decryption key is $(\lambda, \mu)$.

(3) It is greatly recommended to select $p$ and $q$ of equivalent or similar equivalent length in practice to save the cost of calculation time.

## 2.4 Bayesian Optimization

The accuracy of machine learning models not only relies on the algorithm but also on the hyperparameters [62, 63]. Hyperparameters, which directly affect the model's performance and behavior, are settings that aren't learned from the data samples but set by the user before training [64]. Among the various methods available for hyperparameter tuning, Bayesian Optimization (BO) has demonstrated its suitability in XGBoost models [65, 66]. This approach combines prior information from existing parameter selection results with sampling points, continuously updating the probability distribution in the objective function by using Bayesian formula. The prior knowledge obtained from the function distribution would guide the selection of the next parameters, effectively reducing the retrieval times of hyperparameters. In this research, BO is introduced to finetune the hyperparameters of the XGBoost models.

For an aggregation $A = \{(x_1, y_1), (x_2, y_2), ..., (x_t, y_t)\}$ of observations for the previous $t$ step, the observation error $\varepsilon_i$ of the $i$-th step ($i \in [1, t]$) and posterior distribution $D(f|A_{1:i})$ of the objective function f are expressed as follows [67]:

$$\varepsilon_i = y_i - f(x_i) \tag{18}$$

$$D(f|A_{1:i}) \propto D(A_{1:i}|f) \times D(f) \tag{19}$$

where $x_i$ is the hyperparameter of the $i$-th step; $y_i$ represents the observed value for the $i$-th step; $f(x_i)$ is the objective value at the observation point $(x_i, y_i)$; $D(f)$ means the prior distribution of $f$ and $D(A_{1:i}|f)$ is the likelihood distribution of $y_i$.

Gaussian Process (GP) is chosen as the surrogate model in this study to fit data and update the posterior distribution of functions with its advantages of high flexibility and tractability. Compared to traditional methods that aim to mitigate distribution representation, GP differs in its direct modeling of functions, yielding a non-parametric model. One prominent advantage of this approach is its ability to not only simulate any black-box function but also to simulate uncertainty. This quantification of uncertainty is crucial for efficient training processes [67]. GP is the extension of multivariate Gaussian distribution to infinite dimension, which is composed of a mean value function *ave* and a positive-definitive kernel or covariance function *k*.

$$P(x) \sim GP[ave(x), k(x, x')] \tag{20}$$

The RBF kernel function is selected as the covariance function in GP [66]:

$$k(x, x') = \exp\left(\frac{\|x - x'\|^2}{2\sigma^2}\right) \tag{21}$$

where $x$ and $x'$ are the data sample; $\sigma$ is a free parameter set to 1.

Constrained by privacy protection mechanisms, neither the active party nor trusted third party under the FL framework can directly access any of the databases without ownership. Hence, conventional optimization methods, like Bayesian optimization algorithm, are difficult to apply for the proposed FL-XGBoost model. On the other hand, the high communication costs of optimization algorithms under the FL agreement are also a significant obstacle in actual engineering [68]. Based on this, a feasible, secure, and simple Bayesian optimization method has been innovatively introduced for tuning the hyperparameter of the FL-XGBoost models. The optimized aggregated parameters $P_{global}^{BO}$ for the joint models can be obtained by Eq. 21. The optimized parameters for the global model are calculated through proportional aggregation of the optimized parameters based on multiple party local datasets.

$$P_{global}^{BO} = \sum_{i=1}^{N} \frac{n_i}{\sum_{i=1}^{N} n_i} \times P_i^{BO} \tag{22}$$

where $N$ is the number of participants in the joint modeling; $n_i$ is the sample number provided by the participant $i$; $P_i^{BO}$ represents the hyperparameter optimized by the BO algorithm based on the datasets owned by the participant $i$.

## 2.5 Evaluation index

To investigate the benefits of the proposed FL frameworks, several performance metrics are selected as evaluation indexes in this section.

The concept of confusion matrix, which is an essential index to evaluate the results of a binary classification model, is introduced initially. As shown in the left side of Figure 3, it is a summary table with four different combinations: True Positive (TP), False Positive (FP), False Negative (FN), and True Negative (TN) of "Actual Value" and "Predicated Value" [69]. These values in the confusion matrix are used to calculate the following performance metrics.

*Accuracy* of a classifier (ACC) refers to the ratio of correctly classified test samples to the total number of test samples.

$$Accuracy = \frac{TP+TN}{TP+FP+TN+FN} \qquad (23)$$

*Precision* is defined as the ratio of correctly classified test samples with the positive label to the total test sample predicted to be positive.

$$Precision = \frac{TP}{TP+FP} \qquad (24)$$

*Recall* metric represents the ratio of correctly classified positive test samples divided by total number of test samples that are actually positive.

$$Recall = \frac{TP}{TP+FN} \qquad (25)$$

*F1-Score*, a metric designed to work well on imbalanced data, is defined as the harmonic mean of *Precision* and *Recall*, as shown as follows. It is easy to understand from Eq. 25 that an excellent model will obtain a high $F_1$ scores.

$$F_1 = \frac{2(Precision \cdot Recall)}{Precision+Recall} \qquad (26)$$

*False Positive Rate* (FPR) is able to explain what proportion of the negative class got incorrectly classified by the model.

$$FPR = \frac{FP}{FP+FN} \tag{27}$$

The metrics mentioned above are prone to being influenced by varying threshold values across different datasets. To quickly visualize which threshold can yield a better result, the Area Under the Receiver Operating Characteristic (AUC-ROC) curve has been used, as shown on the right side of Figure 3. Receiver Operator Characteristic (ROC) is a probability curve that shows the performance of a classification model at all classification thresholds. The AUC represents the degree or measure of separability, which is used as a summary of the ROC curve. It tells how much the model is capable of distinguishing between classes. Obviously, the higher the AUC, the better the model.

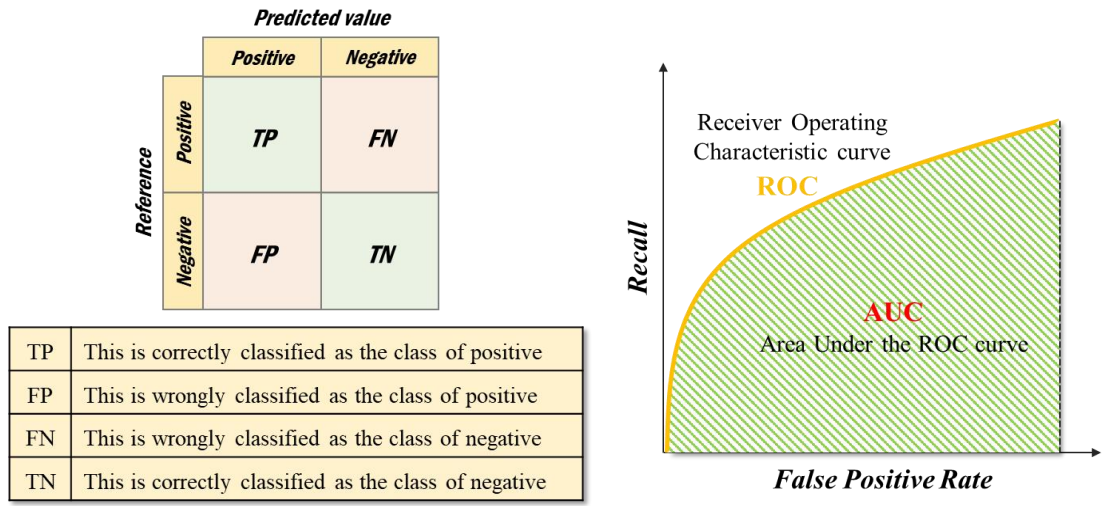

Figure. 3. Illustration for confusion matrix and AUC curve.

## 2.6 XGBoost models under horizontal federated learning frameworks

Following the methodology detailed in Section 2.1, 2.2, and 2.3, a novel privacy-protection FL framework integrating XGBoost models and homomorphic encryption, namely HFL-XGBoost is introduced for expanding the sample size of training data.

The whole training process for the HFL-XGBoost model is illustrated in Protocol

1. The process commences with participants initialize $\{\hat{y}\}_{n_i}$ with random values and obtaining binning points for all features within the HFL framework. Local data samples are then pre-processed accordingly. Each client calculates a local histogram containing gradient $g_i$ and hessian matrix $h_i$ based on their respective datasets. For each tree in the ensemble, the parameter $D$ is defined, signifying the maximum depth of the tree. To mitigate data leakage during histogram exchange, clients introduce a random number to their local histograms before transmitting their unique parameters to the central server (from $g_i$ and $h_i$ transferred to $[gi]$ and $[hi]$). These random numbers cancel each other out through calculation, ensuring that the server can derive global parameters from clients without accessing the original information. The server performs histogram subtractions to identify the best split points. New node histograms are obtained by subtracting the submitted node local histograms from the aggregated node histograms. The best splits points are then broadcasted to clients. Upon receiving and decrypting the new node histograms, clients construct the next layer for the current local model (decision tree), and local data samples are reassigned. Clients compute the leaf nodes of the tree model and update the local histogram. The updated histograms are sent to global server for parameters aggregation after encryption. The process is repeated until the decision tree reaches the required max depth or stop conditions are fulfilled (e.g., maximum tree number or convergences of loss). When the process is completed, a collaborative XGBoost model is established and then sent to every participant. The proposed HFL framework enables petroleum companies to leverage each other's data for training better ML models without compromising the privacy of individual datasets.

---

**Protocol 1**: Horizontal Federated Learning XGBoost model

---

**Input:** $\{X_k\}_{M \times N_k}$ the feature dataset of participant party $k$ ($1 \leq k \leq m$); $\{Y\}_M^k$, the label dataset of participant party $k$; $T$, number of trees; $D$, maximum depth of tree; $B$, number of bins; $\varepsilon$ minimum loss reduction for a split; $\lambda$, L$_2$ regularization term.

**Output:** The $t$-th decision tree for $1 \leq t \leq T$. The main tree structure is sent to each party when the global model is established.

1: **for each party**

2:    Initialized $\{\hat{y}\}_{n_i}$ with random values.

3: **end**

4: **for** $1 \leq t \leq T$ **do**

5:    **for each party**

6:    Party $k$: Calculate the gradient $g_i$ and hessian matrix $h_i$ of local database.

7:    Encrypt $g_i$ into $[g_i]$ and $h_i$ into $[h_i]$.

8:    **end**

**Protocol 1 (Continued).**

9:    **for** $1 \leq d \leq D$ **do**

10:    **for** each tree node on depth $d-1$ **do**

11:    **for** $1 \leq k \leq M$ **do**

12:    **for each party**

13:    Do binning on $I$ and X$_k$, getting the binning result $X_k^{bin}$ and binning boundaries $b_{k,j}$ for $1 \leq j \leq Nk$. Then encrypt the $X_k^{bin}$ into $[X_k^{bin}]$, and record the binning boundaries $b_{k,j}$ to local storage.

14:    Based on $[g_i]$ and $[h_i]$ and $[X_k^{bin}]$, calculate aggregated gradients $[G_k]$ and $[H_k]$. Send $[G_k]$ and $[H_k]$ to the Server.

15:    **end**

16:    Server: Decrypt the aggregated gradients $[G_k]$ and $[H_k]$ for all party, to get $G_k$ and $H_k$. Compute the best split of this tree node and record corresponding information about this node into a lookup table, including the split party, the split feature, the split position and split thresholds.

17:    The best splits points are then broadcasted to each party.

18:    Party $k$: According to the feature ID on party $k$ and the position of split points, for sample vector $I$, determine the left sample space $I_L$. Send $I_L$ to the Server.

19:    Server: Compute $I_R \leftarrow I - I_L$. Split the current tree node into two child nodes to join the node queue, assign $I_L$ and $I_R$ to them respectively.

20:    **end**

21:    **end**

22:    **for** each leaf node u in the tree **do**

**Protocol 1 (Continued).**

Compute the weight $\omega_u \leftarrow -\dfrac{\sum_{i \in I_u} g_i}{\sum_{i \in I_u} h_i + \lambda}$

**end**

Add the new generated decision tree to the model

**end**

**return** the generated model with all trees.

## 2.7 XGBoost models under vertical federated learning frameworks

Unlike HFL, vertical federated learning (VFL) holds promise in safe expansion of data features among various industries. In this section, a VFL method combined with XGBoost, namely VFL-XGBoost, has been illustrated.

Protocol 2 shows the detailed training process of VFL-XGBoost model. For the $t$-th tree within the model, the workflow starts from computation of predicted values $\hat{y}_i\,(1 \leq i \leq M)$ from the preceding $t{-}1$ tree. Subsequently, the first- and second-order gradients, denoted as $g_i$ and $h_i\,(1 \leq i \leq M)$ respectively, undergo computation. These gradients are then subjected to encryption and transmission as $[g_i]$ and $[h_i]\,(1 \leq i \leq M)$ to the cryptographic endpoint. The construction of the tree commences with the creation of the root node $u_0$ and the initialization of the sample space $I_{u_0}$, which encompasses the entire dataset. The iterative node splitting process progresses through each depth. At each depth, the active participant shares its sample space $I$ with all passive participants. Subsequently, each participant $k$ conducts binning on its local feature dataset $X_k$ within the shared sample space $I$. The outcomes of this binning process are encrypted as $[X_k^{bin}]$ and transmitted to the cryptographic endpoint, with the binning boundaries stored locally. Upon receiving the encrypted binning results, the cryptographic endpoint aggregates the gradients based on these results and transmits the aggregated gradients $[G_k]$ and $[H_k]$ to the active participant. The active participant, upon decryption, calculates the information gain for each feature from every participant, determining the optimal split. The active participant then communicates this optimal split to the respective participant, which employs it to update its sample space and initiate the next level of the tree. This process iterates until the tree reaches its defined maximum depth. The weights of the leaf nodes are subsequently computed, and the newly generated tree is incorporated into the ensemble model. Each party only knows the detailed split information of the tree nodes where the split features are provided by the party. By following these steps, multiple parties could collaboratively build a XGBoost model based on the VFL framework without privacy leakage.

**Protocol 2:** Vertical Federated Learning XGBoost model

**Input:** $\{X_k\}_{M \times N_k}$ the feature dataset of participant party $k$ ($1 \leq$ k $\leq$ m); $\{Y\}_M$, the label dataset; $T$, number of trees; $D$, maximum depth of tree; $B$, number of bins; $\varepsilon$ minimum loss reduction for a split; $\lambda$, $L_2$ regularization term.

**Output:** The $t$-th decision tree for $1 \leq t \leq T$. The main tree structure is stored on the active party, while the binning boundaries are stored distributed across all parties.

1:    **for** $1 \leq t \leq T$ **do**

2:        Active party: Initialized and calculate the gradient $g_i$ and hessian matrix $h_i$

3:        Encrypt $g_i$ into $[g_i]$ and $h_i$ into $[h_i]$

4:        Add the root node $u_0$ to the tree, set the sample space $I_{u_0}$

5:        **for** $1 \leq d \leq D$ **do**

6:           **for** each tree node on depth $d-1$ **do**

7:           Active party: For this tree node, send its sample space $I$ to all passive parties

**Protocol 2 (Continued).**

8:        **for** $1 \leq k \leq M$ **do**

9:           Passive Party $k$: Do binning on $I$ and $X_k$, getting the binning result $X_k^{bin}$ and binning boundaries $b_{k,j}$

           **for** $1 \leq j \leq N_k$ **do**

           Then encrypt the $X_k^{bin}$ into $[X_k^{bin}]$, and record the binning boundaries $b_{k,j}$ to local storage.

10:       Passive Party $k$: Based on $[g_i]$ and $[h_i]$ and $[X_k^{bin}]$, calculate aggregated gradients $[G_k]$ and $[H_k]$. Send $[G_k]$ and $[H_k]$ to the active party.

11:     **end**

12:     Active party: Decrypt the aggregated gradients $[G_k]$ and $[H_k]$ for all party, to get $G_k$ and $H_k$. Compute the best split of this tree node and record corresponding information about this node into a lookup table, including the split feature, the split position and split thresholds.

13:     Passive Party $k$: According to the feature ID on party $k$ and the position of split point, for sample vector $I$, determine the left sample space $I_L$. Send $I_L$ to the active party.

14:     Active Party: Compute $I_R \leftarrow I - I_L$. Split the current tree node into two child nodes to join the node queue, assign $I_L$ and $I_R$ to them respectively.

15:   **end**

16:   **end**

17:   **for** each leaf node $u$ in the tree **do**

18:     Compute the weight $\omega_u \leftarrow -\frac{\sum_{i \in I_u} g_i}{\sum_{i \in I_u} h_i + \lambda}$

# 3 Problem Definition

In this section, the reasons behind the existence of data silo in petroleum industry would be analyzed deeply and a FL-based solution, namely FL-XGBoost, is given to overcome this challenge under two different scenarios.

The conventional workflow of gas well exploration is illustrated in the left side of Figure. 4. In the initial phase, two petroleum enterprise (Company A and Company B) commission an exploration institute to gather geological information of their respective reservoirs. Based on the geological features of the two reservoirs, the exploration institute offers tailored development plans, encompassing engineering directives and corresponding technical parameters, to both Company A and Company B. Subsequently, the two asset owners evaluate the feasibility of the proposed plans, taking into account various factors such as technological considerations, economic viability, and societal implications. Once the final construction plans are determined, both companies can initiate the collection of real-time data about the engineering features and productivity of each well in their own reservoirs.

Both geological and operational features of exploration techniques wield a substantial influence on the productivity of undeveloped gas wells. However, as shown on the right side of Figure 4, data barriers pose challenges in the reservoir exploration process. As depicted in Figure.4, the exploration institute hold only the detailed geological features ($G_A$ and $G_B$) of the gas wells in the two reservoirs. Significantly, there is a notable disparity exists between the well development plans, including the operational features $O_{0A}$ and $O_{0B}$, and the actual engineering data ($O_{1A}$ and $O_{1B}$) under the ownership of the Petroleum Company A and B, respectively. The predication objects, productivity ($P_A$ and $P_B$), are exclusively stored in the datasets of the asset owners. Due to the limited availability of features in their respective databases, none of the participating parties can perform EPP tasks or update their existing models in isolation. Furthermore, collaboration barriers often impede data sample expansion between functionally equivalent organizations, like Companies A and B. Concerns related to privacy and the protection of business secrets significantly limit the data availability and diversity, fostering the growth of data silos. Conventional ML solutions heavily rely on the quality and quantity of data, rendering them impractical in scenarios where data is limited [45]. The establishment of accurate ML models becomes challenging, as evident in the literature, without a secure data-sharing agreement between participants due to the existing data barriers.

As previously analyzed, the difficulty in reaching similar open-access agreements is increasing exponentially. Hence, the adoption of FL is urgently needed to address this problem.

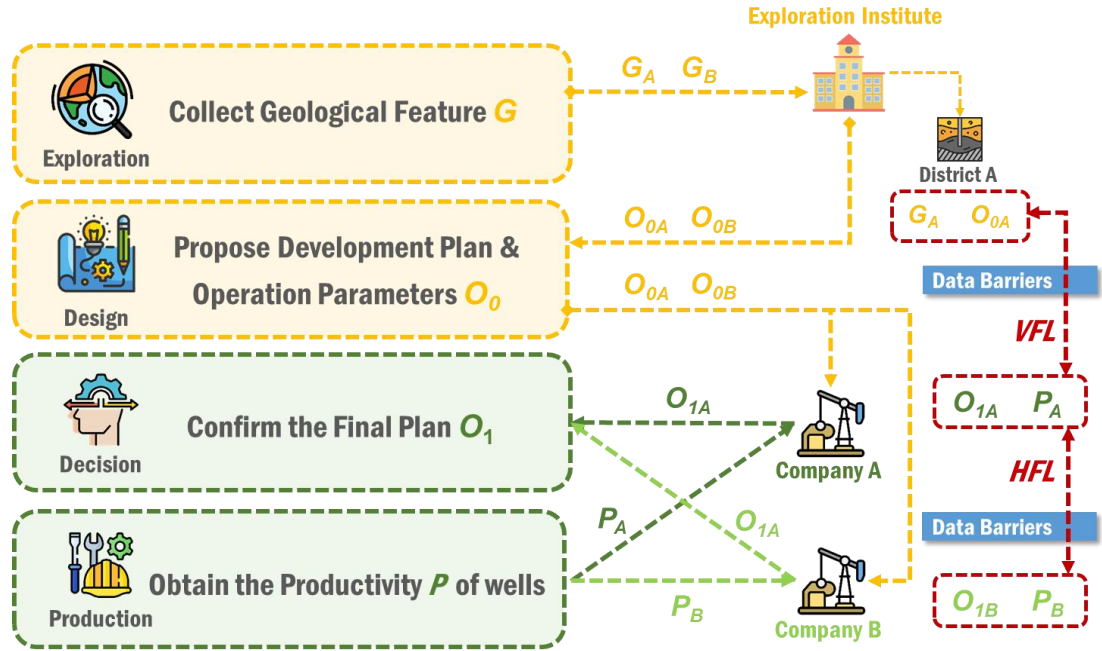

Figure. 4. Data silo in the process of exploration and drilling for gas wells.

## 3.1 Two designed cases of FL-XGBoost frameworks

In the previous section, an exhaustive analysis of the factors contributing to data barriers was presented. Moderately increasing the quantity of data samples can enhance the performance of ML models. Unfortunately, in practice, data samples from gas wells are often limited since the extremely high exploration and drilling costs. Another significant challenge in applying ML algorithms is the scarcity of feature samples for objects. Reservoir exploration involves diverse participants with distinct roles in the upstream, midstream, or downstream sectors of the industrial chain. Consequently, the same ID samples with different features are collected and stored in different institutions, leading to the prevalent issue of data barriers in the petroleum field.

To address these issues, the HFL-XGBoost and VFL-XGBoost models are expected to securely expand sensitive gas well data in terms of both sample and feature quantities. Two scenarios have been designed for the application of the two frameworks.

**Case One**: two petroleum companies, each with its data of own gas wells, collaborate to train a global XGBoost model without exposing data to others. The specific workflow is presented in Figure 5. Oil Company A and B, each holding labeled samples with the same features, are clients of the HFL agreement, while the centralized server, the platform's dominant administrator, is introduced. The server is responsible for collecting, computing, updating and sending the parameters of ML models between clients.

**Case Two**: an oil company and an exploration institute collaborate to build a XGBoost model for predicting the productivity of unconventional reservoirs. The oil company, holding the label $Y$ and the engineering data matrix of petroleum wells in field A and B, takes on the active party role in the scenario. As shown in Figure 6, the active party assumes a dominant position in the VFL framework. Simultaneously, the exploration institute, equipped with only the geological feature matrix of the same wells, acts as the passive party in this cooperation. The passive party aims to build a model to predict the label $Y$ for its own purposes and plays the role of a client in the VFL setting.

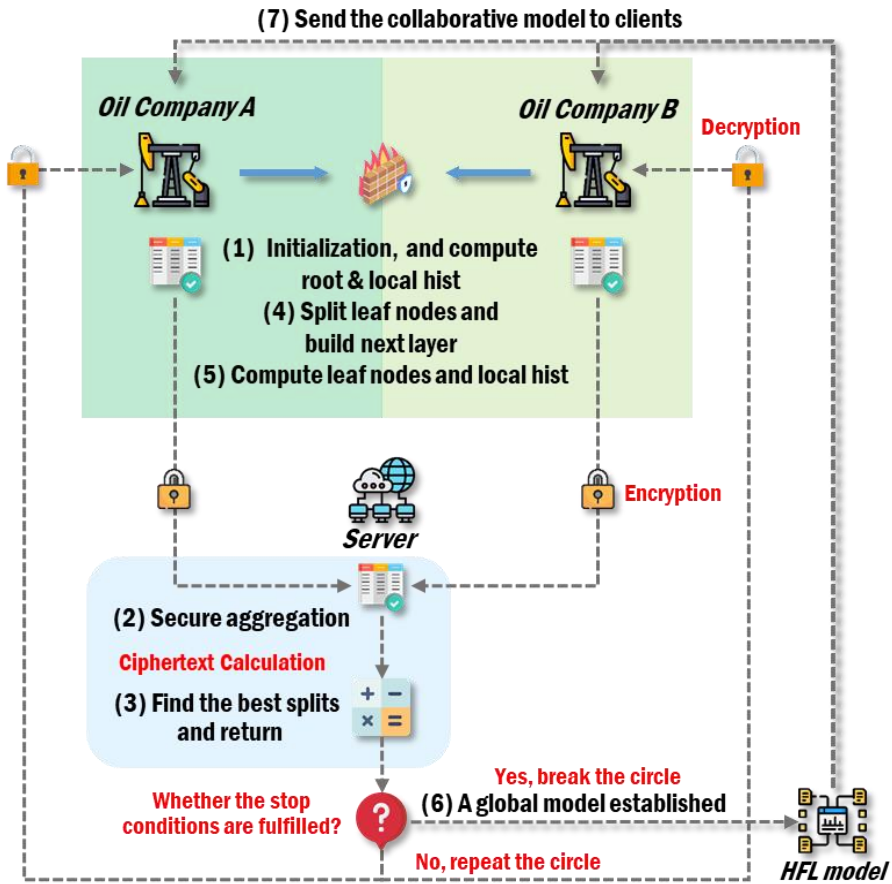

Figure. 5. Workflow of XGBoost algorithm in the HFL framework.

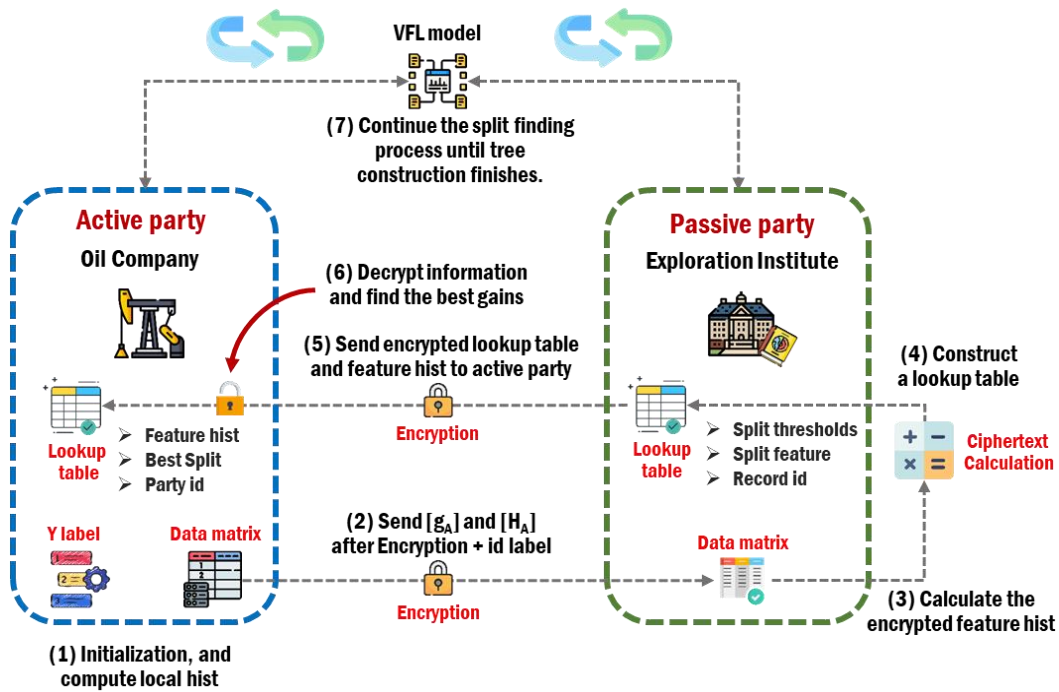

Figure. 6. Workflow of XGBoost algorithm in the VFL framework.

# 4 Experiment and results

## 4.1 Data description and experiment settings

In this section, an effective privacy-preserving classification framework, namely FL-XGBoost is established for the detection of high-yield gas wells, and its benefits are estimated using real-world datasets consisting of a total of 284 gas wells from two shale gas fields, Field A and B, which are located in the Sichuan Basin, China. The target feature in this study is the yield productivity of gas wells and a criterion is introduced for classifying a gas well as high-yield. Based on the essential information provided by the gas field owners, the criterion is set at $2 \times 10^4$ m$^3$ per day. In other words, if the average productivity of the gas well exceeds this threshold, it is defined as a high yield well (positive sample) in this study. Otherwise, it is classified as a low yield well (negative sample). The sample composition of the dataset is shown in Table 1. Raw data undergoes preprocessing before modelling to ensure the completeness of features and values, which is crucial for the accuracy of the models. During this step, features with a high number of missing values are deleted from the dataset. Consequently, 16 operational features and 16 geological parameters are selected as input features for the models. The nomenclature table of the selected featured is provided in Table 2. Besides, more details of the input features are summarized in Appendix A, Table A1 and A2.

To enhance dataset partition certainty, $k$-folds cross-validation, a resampling method has been applied in this research [70]. The preprocessed data is divided into $k$ groups of approximately equal size [71]. In each test round, one group is treated as the testing set, while the remaining $k$-1 groups are used as the training set. This process is repeated $k$ times, with each group serving as the testing set once. The evaluation results from each round are then averaged to estimate the performance of the model.

In this research, $k$ is chosen as 5. In each iteration, 80% of the data is utilized as the training dataset, and the remaining 20% is allocated as the test set. During the modelling and hyperparameter tuning process, 10% of the training data was randomly selected as a validation set. To directly illustrate the gains and costs of the FL framework, a comparison and analysis of the separate model, the federated model, and the centralized model are conducted in the case of HFL and VFL, as summarized in Table 3.

The separate models and centralized models are constructed using Python platform with Scikit-learn ML library [72]. As for federated models, the framework is supported by FATE, an open-source project initiated by Webank's AI Department [73]. The basic settings and parameter selections for the three types of XGBoost models are identical, as listed in Table 4. More specific illustration for the basic settings of XGBoost models can be found in in Appendix B, Table B1. Furthermore, BO algorithm is employed to enhance the performance of the models by tuning hyperparameters. Due to differences in structure and settings between Scikit-learn and FATE, 8 adjustable hyperparameters is chosen to be optimized, as summarized in Table 4. Since FATE platform currently does not support the conventional BO method, which requires global data information, is unavailable restrained by the FL agreement. Therefore, the optimized hyperparameter of FL-XGBoost models are obtained through proportional aggregation of the optimized parameters based on participants' local models, as calculated in Eq.22. This method is referred to aggregated optimization, distinguishing it from the traditional optimization.

Table 1. Data size of 284 shale gas wells in the study case.

| Item | District A | District B | Total |
|---|---|---|---|
| Number of positive samples | 25 (34.72%) | 144 (68.87%) | 171 (60.21%) |
| Number of negative samples | 47 (65.28%) | 66 (31.13%) | 113 (39.79%) |
| Sum | 72 | 212 | 284 |

Table 2. Nomenclature table of the input features in the study case.

Geological parameters

| Item | Units | Symbol |
|---|---|---|
| Total horizontal section length (HSL) | m | $G_1$ |
| HSL in section 'Jiancaogou' | m | $G_2$ |
| HSL in section 1 | m | $G_3$ |
| HSL in section 2 | m | $G_4$ |
| HSL in section 3 | m | $G_5$ |
| HSL in section 4 | m | $G_6$ |
| HSL in section 5 | m | $G_7$ |
| HSL in section 6, 7, 8 and 9 | m | $G_8$ |
| Total HSL in section 1 and 3 | m | $G_9$ |
| Ordinate of the wellhead | - | $G_{10}$ |
| Abscissa of the wellhead | - | $G_{11}$ |
| Middle depth of the well | m | $G_{12}$ |
| Porosity of reservoir | % | $G_{13}$ |
| Total organic carbon of reservoir | % | $G_{14}$ |
| Formation pressure coefficient | - | $G_{15}$ |
| Mean fracture pressure | MPa | $G_{16}$ |

Operational parameters

| Item | Units | Symbol |
|---|---|---|
| Content of slick water in fracturing fluid | $m^3$ | $O_1$ |

| Content of guanidine gum in fracturing fluid | m$^3$ | $O_2$ |
| Total content of liquid in fracturing fluid | m$^3$ | $O_3$ |
| Total content of quartz sand in fracturing fluid | m$^3$ | $O_4$ |

Table 2 (Continued). Nomenclature table of the input features in the study case.

| Operational parameters | | |
| --- | --- | --- |
| Item | Units | Symbol |
| Average content of quartz sand in fracturing fluid | m$^3$ | $O_5$ |
| Average content of quartz sand in sections | m$^3$ | $O_6$ |
| Average content of liquid in sections | m$^3$ | $O_7$ |
| Average content of liquid in clusters | m$^3$ | $O_8$ |
| Content of 30-50 mesh proppant | m$^3$ | $O_9$ |
| Content of 40-70 mesh proppant | m$^3$ | $O_{10}$ |
| Content of 70-140 mesh proppant | m$^3$ | $O_{11}$ |
| Average ratio of quartz sand | - | $O_{12}$ |
| Cluster number of perforations | - | $O_{13}$ |
| Mean pump pressure | MPa | $O_{14}$ |
| Average length of fracturing stages | m | $O_{15}$ |
| Fracturing stages | - | $O_{16}$ |

Table 3. Description of modelling in the study case.

| Modelling Description | Privacy | Generalizable |
| --- | --- | --- |
| **Separate modelling**: Each party independently builds an ML model using its own database | **Yes** | No |
| **Centralized modelling**: All parties could collaborate | No | **Yes** |

with each other by openly data exchange

| | | |
|---|---|---|
| **Federated modelling**: Safe centralized modelling | **Yes** | **Yes** |

Table 4. Basic information for the setting of FL-XGBoost model.

Basic setting (Fixed)

| Item | Details | Item | Details |
|---|---|---|---|
| booster | gbtree | colsample_bylevel | 1 |
| seed | 0 | colsample_bytree | 1 |
| eval_metric | auc | max_delta_step | 0 |
| tree_method | hist | Gamma | 0 |
| random_state | 0 | num_parallel_tree | 1 |
| objective | binary: logistic | | |

Hyperparameter (Optimized by OP algorithm)

| Item | Default value | Adjustment range |
|---|---|---|
| learning_rate | 0.3 | [0.01, 0.5] |
| max_bin | 32 | [8, 512] |
| max_depth | 5 | [0, 10] |
| min_child_weight | 1 | [0, 10] |
| n_estimators | 20 | [20, 100] |
| reg_alpha | 0 | [0,1] |
| reg_lambda | 0 | [0,1] |
| subsample | 1 | [0.01, 1] |

## 4.2 Experiments and results in the case of HFL

In the case of HFL, where two petroleum companies with different labeled samples but the same features collaborate to build a global model while avoiding privacy leakage, the prediction results of different models are presented in Table 5. The aggregated optimized parameters and the direct optimized parameters for the HFL-XGBoost model are collected respectively in Appendix B, Table B2 and Table B3.

According to the results in Table 5, the overall performance of the XGBoost model improves dramatically as the number of sample clusters increases. The federated XGBoost framework outperforms the separate model based on the local dataset of Company A by up to 20.08% in AUC, 26.73% in ACC and 36.98% in F1-Score. For Company B, which contributes nearly 75% of the data samples, the gains from collaborative modeling are still remarkable. AUC, ACC, and F1-Score are 4.55%, 5.67%, and 2.39% higher respectively than the local model. This improvement can be attributed to the larger data volume provided by other participants, which enhances the knowledge available for the ML models. More importantly, the federated XGBoost model achieves an almost perfect trade-off between privacy and accuracy. Calculated by Eq.9, the costs of privacy preservation in AUC, ACC and F1-score are 0.28%, 1.41%, and 1.27% under the proposed FL framework, while the centralized models have a higher average cost of 28.92% (Company A) and 5.19% (Company B) for privacy protection across these evaluation indicators.

Apparently, there was a quick performance boost after hyperparameters tuning, especially for the separate models. FL model received a slight improvement from the safe aggregated optimization, with AUC and F1-Score increasing by 0.54% and 0.34%. In contrast, the centralized XGBoost model with the same setting showed more notable gains in each evaluation index, increasing by an average of 1.78%. Two factors might contribute to this phenomenon. First, the BO module in the FL framework only obtains locally optimal hyperparameters from the selected independent datasets, lacking the complete information about the full data samples. The second reason is that communication loss between encryption and decryption might further aggravate this problem. Hence, it is understandable that the direct BO algorithm can provide a more notable growth for the centralized model, with an average benefit of up to 3.74%, which is higher than the benefit from the aggregated BO. Although BO methods effectively reduce the privacy protection costs of centralized modeling, they still remain twice as high as those of FL models. The average privacy-care costs for the three evaluating indicators of FL models are 0.95% for AUC, 3.90% for ACC, and 2.59% for F1-score.

From the standpoint of the company A and B in the HFL case, Figure 7 illustrates the benefits provided by the different methods in detail and their privacy preservation costs. FL techniques distinctly diminish privacy preservation costs, yielding a mere 0.28% cost on AUC, 1.41% on ACC, and 1.27% on F1-score in the case of HFL. For participants with a small sample dataset, like company A, insufficient data volume is the primary constraint for training models, and embracing collaboration is a wise choice to achieve more accurate prediction results. Besides, centralized modeling incurs significantly higher privacy protection costs than contributors with larger datasets, especially concerning F1-score. There is a decrease in privacy preservation costs by 0.72% on AUC, 3.92% on ACC, and 11.61% on F1-score after hyperparameter tuning.

As shown in Figure 7 (b), the increments in AUC and ACC are obvious for Company B with the 25% increase in data samples, while the improvement of F1-Score is more sensitive to the hyperparameter variation. Hence, selecting appropriate parameters is a prudent solution to effectively enhance the F1-Score of the XGBoost algorithm. Intriguingly, the AUC of the centralized model for Company B decreased by 1.5% after Bayesian Optimization, while ACC and F1-score increased by 1.32% and 2.42%, respectively. This phenomenon is attributed to the great improvement from parameter adjustment in AUC and F1-score for the centralized model compared to the separate model built by Company B. Since direct data access is prohibited under the FL agreement, FL models optimized using the aggregation BO algorithm yield suboptimal outcomes. resulting in a larger increase in the privacy protection cost of FL models after adopting hyperparameter optimization in HFL.

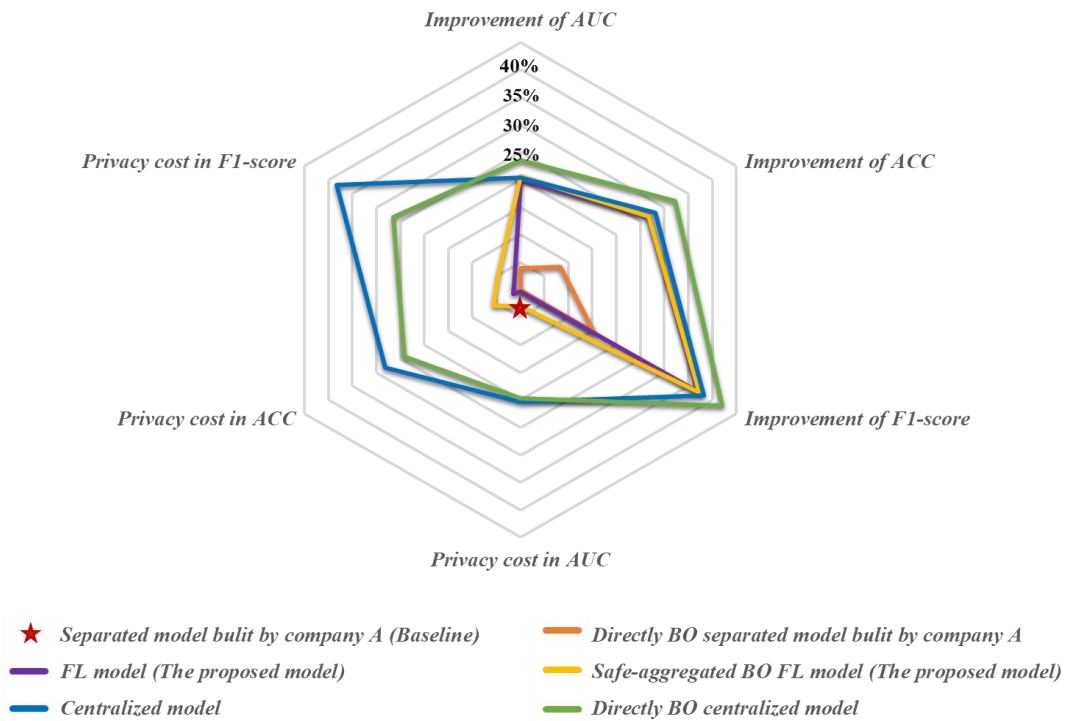

(a) Take separate model built by the company A as the baseline.

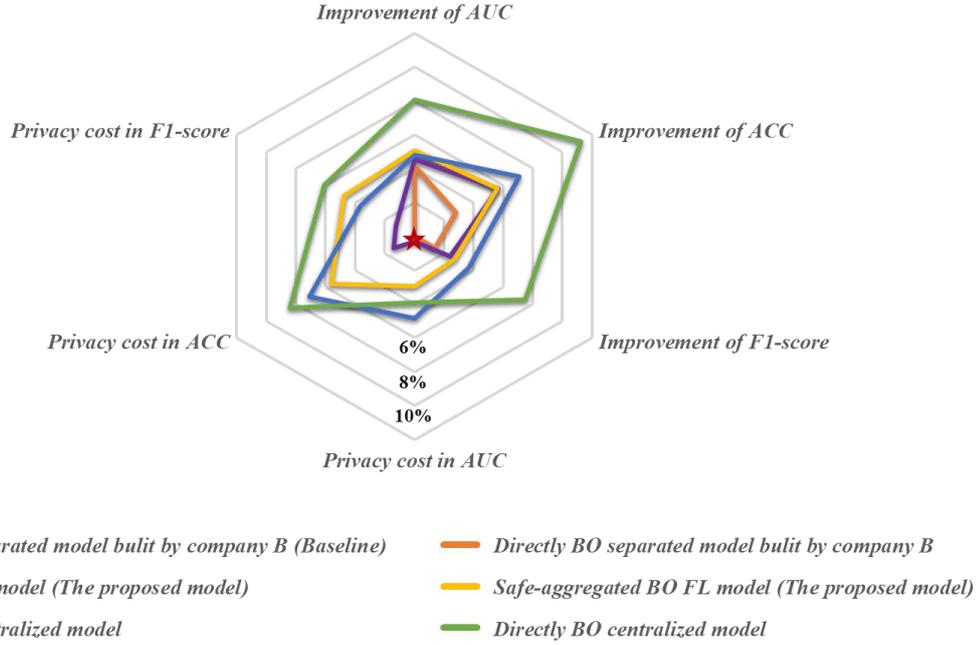

(b) Take separate model built by the oil company B as the baseline.

Figure. 7. The benefits brought from collaborative modelling and hyperparameter

optimization in the case of HFL.

Table 5. The comparison results in the case of HFL.

|  | Item | Privacy | AUC | ACC | F1-Score |
|---|---|---|---|---|---|
| Separate model | Company A | Safe | 69.25% | 56.03% | 48.49% |
|  | Direct BO | Safe | 73.19% | 64.11% | 63.94% |
|  | Company B | Safe | 84.78% | 77.09% | 83.08% |
|  | Direct BO | Safe | 88.93% | 79.93% | 84.50% |
| Federated model | Original | Safe | 89.33% | 82.76% | 85.47% |
|  | Aggregated BO | Safe | 89.87% | 82.73% | 85.81% |
| Centralized model | Original | Not safe | 89.61% | 84.17% | 86.74% |
|  | Aggregated BO | Not safe | 90.82% | 86.63% | 88.40% |
|  | Direct BO | Not safe | 92.83% | 88.33% | 90.58% |

## 4.3 Experiments and results in the case of VFL

Under the VFL agreement, the active party, which possesses the label samples (Y label), can engage the suitable passive party in a collaborative effort to build a model through encrypted communication. In this scenario, the prediction results of various XGBoost models are collected in Table 6. Additionally, the aggregated optimized parameters and the direct optimized parameters for the models are summarized respectively in Appendix B, Table B4 and Table B5.

As outlined in Table 6, the performance of the XGBoost model notably improves with the increase in data dimension, especially concerning AUC. Owning to the stronger correlation between well productivity and the surrounding geological characteristics, the accuracy of the separate model established by the exploration institute surpasses that of the model based solely on operational data of the wells. When the exploration department adopt the FL framework, the AUC rise by 6%, which is twice the increment seen in ACC and F1-score. For the owner of operational data, the benefits derived from the joint FL model are significantly higher, with an improvement of 16.72% in AUC, 7.78% in ACC, and 5.71% in F1-score. In the scenario of local modelling, parameter optimization more readily enhances AUC, particularly for participants with operational data. But whether using the direct BO method or the aggregated BO method, both approaches result in more substantial improvements in ACC and F1-score for collaborative models.

The gains from the different ML models and the privacy-preserving costs are illustrated detailly in the case of VFL. In Figure 8 (a), the most substantial enhancement in AUC of the model based on the oil company exceeds 15% when geological features from the exploration institute are introduced to the training datasets. In Figure 8 (b), operational features lead to a 5.91% improvement in AUC of the separate model within the VFL framework, while the growth in ACC and F1-Score is comparatively limited. One the other hand, both the FL model and centralized model experience satisfactory advancements from the aggregated BO method, especially for ACC, which increases by 3.19% for the FL model and 2.46% for the centralized approach.

In the case of VFL, the privacy-preserving cost of FL models remains significantly lower than that of centralized models, whether for the Oil Company or the Exploration Institutes. Specifically, the FL models exhibit a 5.78% lower privacy protection cost in terms of AUC, a 3.53% lower cost in terms of ACC, and a 3.05% lower cost in terms of F1-score. Although this advantage may slightly diminish after parameter optimization, it still maintains a notable edge compared to the centralized models. Except for the privacy protection cost of the centralized model for the Oil Company, which experienced a 1.89% decrease in AUC after optimization, the cost of all other models showed increments in all metrics.

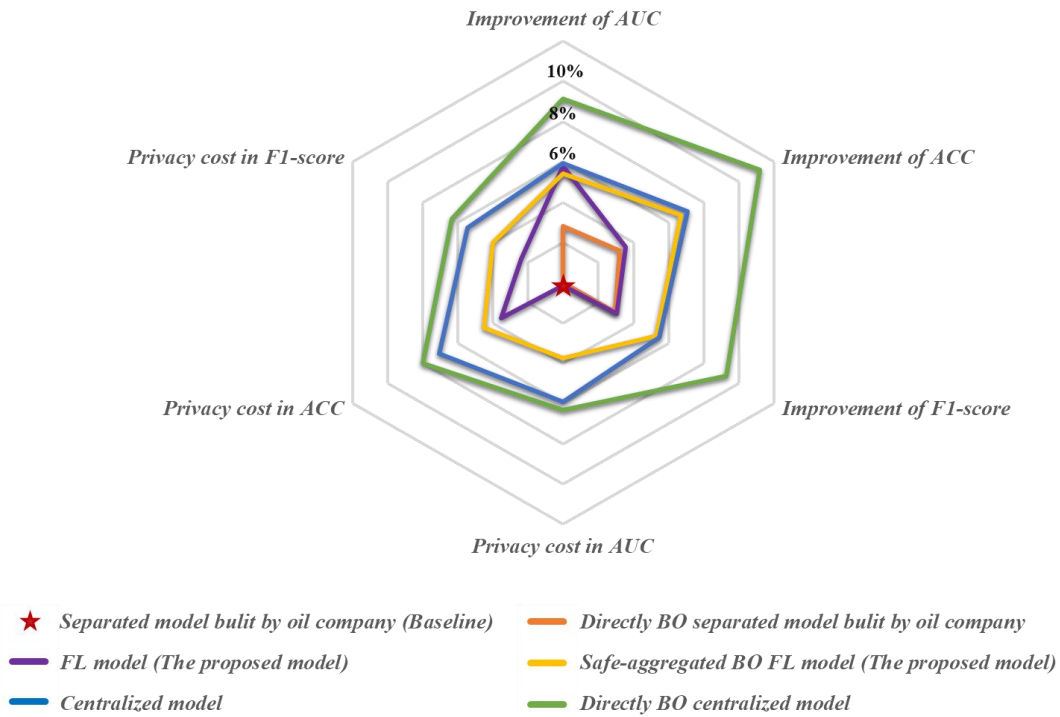

(a) Take separated model built by the oil company as the baseline.

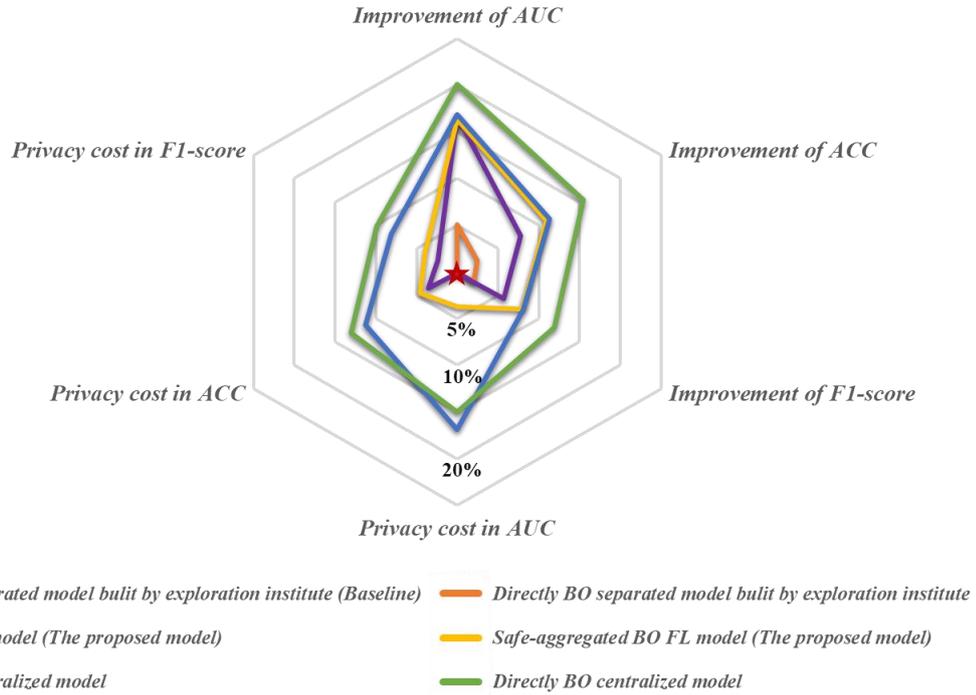

(b) Take separated model built by the exploration institute as the baseline.

Figure. 8. The benefits brought from collaborative modelling and hyperparameter optimization in the case of VFL.

Table 6. The comparison results in the case of VFL.

| Item | | Evaluation index | | | |
|---|---|---|---|---|---|
| | | Privacy | AUC | ACC | F1-Score |
| Separate model | Oil Company | Safe | 72.76% | 72.86% | 78.63% |
| | Direct BO | Safe | 77.87% | 75.33% | 80.72% |
| | Exploration Institute | Safe | 83.70% | 77.11% | 81.29% |
| | Direct BO | Safe | 86.49% | 80.32% | 84.25% |
| Federated model | Original | Safe | 89.48% | 80.64% | 84.34% |
| | Aggregated BO | Safe | 89.10% | 83.82% | 86.57% |
| Centralized model | Original | Not safe | 89.61% | 84.17% | 86.74% |
| | Aggregated BO | Not safe | 90.82% | 86.63% | 88.40% |

| | | | | |
|---|---|---|---|---|
| Direct BO | Not safe | 92.83% | 88.33% | 90.58% |

# 5 Discussion

The practical implementation of ML algorithms in energy fields is impeded by two prominent challenges: privacy protection and data collection. To address this issue, this research proposes a novel privacy-protection framework to overcome data barriers between agencies and companies, facilitating the application of ML methods in energy fields. XGBoost model was chosen to address a classical binary classification problem under the proposed FL framework.

As analyzed in the previous section, the evaluation and comparison of the separate models, centralized models and FL models were executed based on the actual dataset. FL agreements empower participants to leverage data securely and derive collaboration benefits whether in the case of HFL or VFL. FL models manifest higher accuracy and superior generalization ability over the separate models. Simultaneously, the privacy persevering costs associated with FL models are significantly lower than that of the centralized models. Bayesian Optimization emerges as a viable solution for tuning the hyperparameters of the XGBoost models. Following the application of the Bayesian Optimization method to adjust the hyperparameters of the XGBoost models, the performance of all models has been improved to varying degrees. Constrained by the privacy protection agreement, hyperparameter tuning has been identified as a critical issue for the FL models. The safe-aggregation Bayesian optimization is effective and maneuverable in improving the accuracy of XGBoost models.

To sum up, the experimental results successfully demonstrate the feasibility of the FL framework in the geonergy sector. As the starting point of this work, there are still certain issues that need to be discussed.

## 5.1 Training cost

The training cost of ML models, encompassing computational, personnel expenses, and storage, is pivotal for efficient model development. In this research, an AMD Ryzen 9 7950X 16-Core Processor was utilized for the training, optimization, and prediction processes. Based on the results in Section 4, secure data collaborations and satisfactory accuracy under the FL frameworks derive from intricate ciphertext computations and frequent encryption-decryption processes. However, these processes notably escalate the training costs of FL models, as evidenced in Table 7 and Table 8.

From Table 7 and Table 8, the time consumptions of the separate and centralized models basically maintained within 1 second, while the training costs of FL models are close to 7 minutes. Fortunately, the data samples in energy industry, especially in geoenergy sector, are usually insufficient. Hence, while FL frameworks must carry high computational burdens, the time consumption is generally acceptable in practical applications. Additionally, real-time computation is unnecessary in the vast majority of cases within the energy fields.

To achieve effective modelling under the FL framework, several strategies can be implemented. Firstly, optimize the communication mechanisms in the FL process, which can be achieved through techniques such as data compression, differential privacy, or employing more efficient communication protocols [74, 75]. Secondly, selectively uploading updates that significantly impact the global model can help to prioritize the transmission of relevant information [76, 77]. Participants can evaluate the importance of their local updates based on criteria such as their impact on the model's performance or their relevance to specific tasks. Besides, employing techniques such as model pruning and quantization can reduce the complexity of the model without sacrificing performance, thereby decreasing both computational and communication overheads [78, 79]. Finally, adopting incremental learning approaches involves updating the model gradually over time, rather than retraining it from scratch with each iteration. This incremental approach can help to accelerate convergence, especially in scenarios where the underlying data distribution may change gradually over time [80].

Therefore, the efficiency of the proposed FL-XGBoost framework is expected to be overcome. It stands as a pivotal research focus for our future endeavors that build a practical, efficient, privacy-persevering FL framework in the energy field.

Table 7. Comparison results of training and optimization time consumption for various models in the case of HFL (Unit: Second).

|  | Item | Total time consumption |
| --- | --- | --- |
| Separate model | Company A | 0.04 |
|  | Company B | 0.10 |
| Federated model | Original | 391.56 |
| Centralized model | Original | 0.13 |

Table 8. Comparison results of training and optimization time consumption for various models in the case of VFL (Unit: Second).

|  | Item | Time consumption |
| --- | --- | --- |
| Separate model | Oil Company | 0.06 |
|  | Exploration Institute | 0.05 |
| Federated model | Original | 399.56 |
| Centralized model | Original | 0.13 |

## 5.2 Ethical Consideration

Considering the sensitive data in energy fields, FL presents an excellent solution by achieving data availability while preserving data invisibility, thereby overcoming data barriers between agencies and companies at the lowest cost. In the two designed cases, expansions in data volume and features were successfully achieved without compromising the confidentiality of the original data. But, ethical considerations associated with FL should be carefully addressed.

(1) *Transparency*: Federated learning models may lack transparency, making it difficult for participants to understand how decisions are made or identify potential biases in the collaborative models. Hence, even if a more superior predictive model is obtained based on FL technology, the results hardly provide some specific and reliable advice to each participant. Ensuring the transparency and accountability of algorithms is crucial for addressing ethical issues and building trust in FL systems.

(2) *Dishonesty*: A comprehensive management system is needed to regulate FL modelling participants. During the data-share collaboration, dishonest participants may attempt to tamper with the global model or inject malicious code by forging data. These actions not only compromise the security and reliability of the collaborative models but also undermine the trust among participants. So, clear guidelines and protocols should be established to govern the behavior of participants and outline the consequences of engaging in fraudulent activities.

(3) *Algorithm bias*: Algorithm bias cannot be ignored, as the training data for each collaboration may come from different datasets. The active modeling party or trusted third-party may selectively choose joint modeling participants in a biased manner, leading to data bias and resulting in unfair or inaccurate results from the model. It is essential to ensure a fair and representative selection of participants in the FL process to minimize algorithm bias.

# 6 Conclusion

This study comprehensively investigates the significance and feasibility of deploying the FL-XGBoost model for high-yield gas well classification in the geoenergy fields, emphasizing the evaluation of FL framework benefits in two scenarios. The key findings are summarized as follows:

- The proposed FL models effectively enhance data samples and features, ensuring data privacy and significantly the performance of the XGBoost models.

- Whether under HFL or VFL scenarios, the proposed FL-XGBoost model achieves an exceptional balance between privacy and accuracy.

- Constrained by the privacy protection agreement within the FL framework, the employed aggregated Bayesian Optimization approach demonstrates simplicity and effectiveness in improving the accuracy of the XGBoost models.

In summary, FL agreements empower secure and collaborative data utilization, exhibiting higher accuracy and superior generalization ability over individual models. Notably, privacy-preserving costs associated with FL models are significantly lower than those of centralized models. As the information technology landscape evolves, Federated Learning holds promise as a prospective solution for addressing expanding data barriers within the geoenergy sector.

## CRediT authorship contribution statement

**Weike Peng**: Data curation, Investigation, Methodology, Software,

Writing–original draft. **Jiaxin Gao**: Methodology, Data curation, Software. **Yuntian Chen**: Methodology, Project administration, Funding acquisition, Supervision, Writing–review & editing. **Shengwei Wang**: Project administration, Supervision, Writing–review.

# Declaration of Competing Interest

The authors declare that they have no known competing financial interests or personal relationships that could have appeared to influence the work reported in this paper.

# Acknowledgments

This work was supported by the National Natural Science Foundation of China (Grant No. 62106116), China Meteorological Administration Climate Change Special Program (CMA-CCSP) under Grant QBZ202316, Natural Science Foundation of Ningbo of China (No. 2023J027), the High Performance Computing Centers at Eastern Institute of Technology, Ningbo, and Ningbo Institute of Digital Twin.

# Appendix A. Description of the dataset

The two fields are located in the eastern fold belt of the Sichuan Basin, with a relatively higher structural position. Certain characteristics define Field A: a gentle dip (10 to 20°) in the southwest and a steep dip (20 to 30°) in the northeast. The reservoir depth within the Field A exceeds 3,200 m, surpassing that of the main part of the Field B by 500 to 900 m. Positioned to the west of the Field A, the Field B boasts a relatively smoother attitude of the main strata.

Table A1. Input features of 72 shale gas wells in the A district.

| Item | Unit | Average | Median | Range |
|------|------|---------|--------|-------|
| $G_1$ | m | 1570.38 | 1554.00 | [895.63, 2163.00] |
| $G_2$ | m | 17.23 | 0.00 | [0.00, 173.40] |
| $G_3$ | m | 243.23 | 178.15 | [0.00, 1036.00] |
| $G_4$ | m | 102.49 | 72.30 | [0.00, 417.00] |
| $G_5$ | m | 1019.13 | 1057.25 | [0.00, 1788.00] |
| $G_6$ | m | 127.79 | 51.75 | [0.00, 1362.00] |
| $G_7$ | m | 43.23 | 0.00 | [0.00, 912.00] |
| $G_8$ | m | 17.28 | 0.00 | [0.00, 836.00] |
| $G_9$ | m | 1262.36 | 1328.25 | [0.00, 1975.60] |
| $G_{10}$ | - | 3291330.10 | 3291291.00 | [3283866.00, 3297271.10] |
| $G_{11}$ | - | 18738860.64 | 18739164.65 | [18731627.60, 18747507.10] |
| $G_{12}$ | m | 3463.95 | 3530.85 | [2399.20, 4345.74] |
| $G_{13}$ | % | 4.42 | 4.44 | [3.57, 5.07] |
| $G_{14}$ | % | 3.93 | 3.95 | [2.46, 4.89] |
| $G_{15}$ | - | 1.60 | 1.60 | [1.30, 1.90] |

Table A1 (Continued). Input features of 72 shale gas wells in the A district.

| Item | Unit | Average | Median | Range |
|------|------|---------|--------|-------|
| $G_{16}$ | MPa | 79.23 | 78.16 | [65.45, 100.77] |
| $O_1$ | m$^3$ | 35471.40 | 34738.04 | [18001.00, 60115.00] |
| $O_2$ | m$^3$ | 2897.63 | 2992.30 | [223.40, 6181.68] |
| $O_3$ | m$^3$ | 39897.46 | 38964.49 | [20013.58, 65022.80] |
| $O_4$ | m$^3$ | 1385.30 | 1254.15 | [604.60, 2682.60] |
| $O_5$ | m$^3$ | 68.05 | 61.37 | [40.84, 107.30] |
| $O_6$ | m$^3$ | 474.61 | 628.20 | [9.18, 920.02] |
| $O_7$ | m$^3$ | 1956.57 | 1933.50 | [1229.13, 2335.37] |
| $O_8$ | m$^3$ | 1535.71 | 1887.98 | [187.98, 2196.65] |
| $O_9$ | m$^3$ | 62.42 | 24.15 | [0.00, 968.00] |
| $O_{10}$ | m$^3$ | 917.03 | 773.37 | [214.30, 1861.60] |
| $O_{11}$ | m$^3$ | 405.84 | 353.83 | [108.90, 921.10] |
| $O_{12}$ | - | 6.47 | 6.49 | [5.18, 8.55] |
| $O_{13}$ | - | 76.74 | 60.00 | [28.00, 255.00] |
| $O_{14}$ | MPa | 45.22 | 44.74 | [30.63, 57.74] |
| $O_{15}$ | m | 76.93 | 78.72 | [57.73, 92.72] |
| $O_{16}$ | - | 20.13 | 20.00 | [10.00, 30.00] |

Table A2. Data characteristics of 212 shale gas wells in the B district.

| Item | Unit | Average | Median | Range |
|------|------|---------|--------|-------|
| $G_1$ | m | 1468.05 | 1500 | [786.00, 2203.00] |
| $G_2$ | m | 28.63 | 0.00 | [0.00, 551.50] |
| $G_3$ | m | 386.18 | 306.00 | [0.00, 1748.00] |
| $G_4$ | m | 89.30 | 50.00 | [0.00, 666.00] |
| $G_5$ | m | 743.99 | 726.00 | [0.00, 1613.00] |
| $G_6$ | m | 152.25 | 39.00 | [0.00, 1303.00] |
| $G_7$ | m | 54.09 | 0.00 | [0.00, 757.00] |
| $G_8$ | m | 18.47 | 0.00 | [0.00, 471.00] |
| $G_9$ | m | 1111.93 | 1216.60 | [0.00, 1867.20] |
| $G_{10}$ | - | 3284124.81 | 3284647.66 | [3272450.10, 3296023.60] |
| $G_{11}$ | - | 18743263.60 | 18743721.90 | [18732341.10, 18751582.80] |
| $G_{12}$ | m | 2728.33 | 2646.00 | [1278.20, 5645.00] |
| $G_{13}$ | % | 4.71 | 4.71 | [2.93, 6.37] |
| $G_{14}$ | % | 3.59 | 3.62 | [2.00, 5.03] |
| $G_{15}$ | - | 1.42 | 1.46 | [0.98, 1.69] |
| $G_{16}$ | MPa | 72.65 | 73.17 | [40.02, 88.87] |
| $O_1$ | $m^3$ | 31159.44 | 31170.63 | [3683.58, 57188.90] |
| $O_2$ | $m^3$ | 2611.47 | 2391.16 | [129.40, 7914.90] |
| $O_3$ | $m^3$ | 35173.27 | 34940.60 | [3840.13, 59781.90] |
| $O_4$ | $m^3$ | 966.16 | 988.81 | [111.90, 1786.00] |
| $O_5$ | $m^3$ | 50.27 | 50.65 | [31.22, 79.45] |
| $O_6$ | $m^3$ | 20.06 | 20.06 | [12.54, 29.79] |
| $O_7$ | $m^3$ | 1803.70 | 1820.16 | [1240.64, 2178.36] |

Table A2 (Continued). Data characteristics of 212 shale gas wells in the B district.

| | | | | |
|---|---|---|---|---|
| $O_8$ | m³ | 706.50 | 701.00 | [489.51, 1009.85] |
| $O_9$ | m³ | 41.55 | 38.80 | [0.00, 241.90] |
| $O_{10}$ | m³ | 696.12 | 709.20 | [87.30 1273.80] |
| $O_{11}$ | m³ | 228.49 | 215.20 | [19.60, 438.50] |
| $O_{12}$ | - | 7.44 | 7.28 | [4.81, 23.70] |
| $O_{13}$ | - | 48.37 | 49.00 | [5.00 75.00] |
| $O_{14}$ | MPa | 35.69 | 34.91 | [16.08, 64.48] |
| $O_{15}$ | m | 74.07 | 75.00 | [36.00, 97.00] |
| $O_{16}$ | - | 18.67 | 19.00 | [2.00, 29.00] |

# Appendix B. Illustration of hyperparameters of FL-XGBoost model

Table B1. Illustration for the setting parameters of FL-XGBoost model.

| Parameter | Description |
|---|---|
| colsample_bylevel | The subsample ratio of columns for each level |
| colsample_bytree | The subsample ratio of columns when constructing each tree |
| gamma | Minimum loss reduction required to make a further partition on a leaf node of the tree |
| learning_rate | Step size shrinkage used in update to prevents overfitting |
| max_bin | Maximum number of bins to bucket continuous features |
| max_delta_step | Maximum delta step of each leaf output |
| max_depth | Maximum depth of a tree |
| min_child_weight | Minimum sum of instance weight needed in a child. |
| n_estimators | The number of decision trees. |
| subsample | Subsample ratio of the training instances. |
| booster | The kind of booster used in XGBoost |
| objective | Learning task and objective |
| reg_alpha | L1 regularization term on weights |
| reg_lambda | L2 regularization term on weights |
| eval_metric | Evaluation metrics for testing dataset |
| tree_method | The tree construction algorithm used in XGBoost |
| seed | Random number seed |

Table B2. The optimized aggregation parameters applied for the FL-XGBoost model and the global model in the case of HFL.

| Item | 1 | 2 | 3 | 4 | 5 |
|------|------|------|------|------|------|
| learning_rate | 0.5 | 0.47 | 0.40 | 0.43 | 0.5 |
| max_bin | 343 | 27 | 8 | 16 | 8 |
| max_depth | 4 | 4 | 10 | 6 | 9 |
| min_child_weight | 0.34 | 0.34 | 0.66 | 1.66 | 2.32 |
| n_estimators | 30 | 28.3 | 23.20 | 29 | 28.64 |
| reg_alpha | 0.66 | 0.66 | 0.66 | 0.66 | 0.66 |
| reg_lambda | 0.34 | 0.34 | 0.34 | 0 | 0 |
| Subsample | 0.85 | 1.00 | 1.00 | 0.78 | 0.74 |

Table B3. The optimized parameters applied for the global model in the case of HFL.

| Item | 1 | 2 | 3 | 4 | 5 |
|------|------|------|------|------|------|
| learning_rate | 0.5 | 0.5 | 0.4 | 0.5 | 0.2011 |
| max_bin | 8 | 8 | 8 | 512 | 188 |
| max_depth | 10 | 1 | 8 | 10 | 10 |
| min_child_weight | 1 | 0 | 0.66 | 3 | 3 |
| n_estimators | 0.5 | 0 | 0.66 | 1 | 1 |
| reg_alpha | 0 | 1 | 0.3 | 0.8 | 0 |
| reg_lambda | 0.7273 | 1 | 1 | 1 | 1 |
| Subsample | 30 | 30 | 20 | 30 | 30 |

Table B4. The optimized aggregation parameters applied for the FL-XGBoost model and the global model in the case of VFL.

| Item | 1 | 2 | 3 | 4 | 5 |
|---|---|---|---|---|---|
| learning_rate | 0.12 | 0.50 | 0.48 | 0.50 | 0.15 |
| max_bin | 260 | 271 | 76 | 264 | 207 |
| max_depth | 9 | 6 | 4 | 6 | 10 |
| min_child_weight | 1.5 | 1 | 1 | 1.5 | 0 |
| n_estimators | 30 | 20 | 20 | 23.5 | 28 |
| reg_alpha | 0.5 | 0.5 | 0 | 0 | 1 |
| reg_lambda | 0.5 | 1 | 0.5 | 0.5 | 0.5 |
| subsample | 0.93 | 0.84 | 0.85 | 0.71 | 0.58 |

Table B5. The optimized parameters applied for the global model in the case of VFL.

| Item | 1 | 2 | 3 | 4 | 5 |
|---|---|---|---|---|---|
| learning_rate | 0.1304 | 0.1407 | 0.2900 | 0.1772 | 0.2655 |
| max_bin | 512 | 8 | 128 | 52 | 45 |
| max_depth | 7 | 10 | 8 | 8 | 7 |
| min_child_weight | 2 | 1 | 2 | 4 | 1 |
| n_estimators | 0 | 0 | 0 | 1 | 0 |
| reg_alpha | 1 | 0 | 0 | 0 | 0 |
| reg_lambda | 0.5 | 0.3042 | 0.9897 | 0.6566 | 0.51 |
| subsample | 30 | 30 | 28 | 30 | 15 |